\documentclass{article}

\PassOptionsToPackage{numbers}{natbib}
\usepackage[final]{neurips_2021_ml4ps}

\usepackage[utf8]{inputenc} 
\usepackage[T1]{fontenc}    
\usepackage{hyperref}       
\usepackage{url}            
\usepackage{booktabs}       
\usepackage{amsfonts}       
\usepackage{nicefrac}       
\usepackage{microtype}      
\usepackage{xcolor}         
\usepackage{graphicx}
\usepackage{amssymb}
\usepackage{amsmath}
\usepackage{subfigure}
\usepackage{longtable}

\usepackage{xcolor}

\title{Neural network is heterogeneous: Phase matters more}

\author{%
  Yuqi Nie\\
  Department of Electrical and Computer Engineering\\
  Princeton University\\
  \texttt{ynie@princeton.edu} \\
  \And
  Hui Yuan \\
  Department of Electrical and Computer Engineering\\
  Princeton University\\
  \texttt{huiyuan@princeton.edu} \\
}

\begin{document}

\maketitle

\begin{abstract}
  We find a heterogeneity in both complex and real valued neural networks with the insight from wave optics, claiming a much more important role of phase than its amplitude counterpart in the weight matrix. In complex-valued neural networks, we show that among different types of pruning, the weight matrix with only phase information preserved achieves the best accuracy, which holds robustly under various settings of depth and width. The conclusion can be generalized to real-valued neural networks, where signs take the place of phases. These inspiring findings enrich the techniques of network pruning and binary computation.
\end{abstract}

\section{Introduction}

Neural network has been one of the most attractive topics in the past decade and tremendous progress has been achieved in real-valued neural networks. However, the pace of making progress in complex-valued neural networks is relatively lagged even though people have started exploring in the field since last century. One of the core challenges is the standard complex derivatives cannot be used in designing algorithms to optimize complex-valued neural networks \citep{Simone2018}. Besides, there are non-intuitive analytical properties of complex algebra preventing us from simply replacing the real values in neural networks by complex values. Fortunately, countermeasures are coming up. Examples of them are using Wirtinger’s calculus or optimizing the real and imaginary components separately \citep{popa2017}. Another issue to consider in complex-valued neural networks is adapting the activation function for complex values. To solve it, traditional activation functions have been expanded to complex domain. For $ReLU$ function, it has been transformed into $zReLU$, $\mathbb{C}ReLU$ and $modReLU$ \citep{Chiheb2018}. 

After providing remedies to such challenges, it becomes possible to train complex-valued neural networks successfully. Unfortunately, most of the results have shown that complex-valued neural networks achieve equal or slightly worse performance in comparison to real-valued networks for a range of real-valued classification tasks \citep{Monning2018,Anderson2017}, which might result in the less attention that complex-valued neural networks are attracting than real ones, since complex value requires more storage space and computation power. 

However, we can find some non-trivial intuition from optical physics. Wave optics and neural networks share some similarities, and researchers claimed an analogy existing between those two in recent literature \citep{Hughes2019}. We discuss some of the details in Appendix \ref{app:1}. There are a line of influential papers focusing on the wave phases in optical process. For instance, Zernike's phase contrast method won the Nobel Prize in Physics in 1953 and it is still widely used till today \citep{zernike1955,richards1956,ruane2020}. All these influential works are in a great agreement that phase is extremely important in waves, even more important than amplitude is.

Motivated by this phenomenon, it is natural to ask whether phase still plays an important role in complex-valued neural networks. Our ways to investigate the importance of phase over amplitude are motivated by network pruning, where only "important" connections between layers are reserved for further fine tuning to reduce computation and storage cost. A line of empirical literature shows that pruning can reduce the parameter counts by over $90\%$ without hurting the predicting accuracy, see \citep{lecun1989optimal, han2015learning, li2016pruning} for instance. 

In this paper we claim that phase plays a much more important role than its amplitude counterpart, which reveals the heterogeneity in neural networks. This heterogeneity means (but is not limited to) the phenomenon that more information about the neural network are stored in the phase subset of the whole parameter space. We have run experiments with several depth and width settings of complex-valued neural network, which presents the robustness of our conclusion in different scenarios.  We also expand our result to the real-valued neural networks, where the signs of the matrices' elements take the place of the phases, and we are surprised to find the result in the previous section still holds. Finally we summarize our findings and suggest directions for future research. Our code is available in the repository \href{https://github.com/machinelearninggo/Phase-in-NN.git}{\textcolor{blue}{Phase-in-NN}}
on GitHub.

\section{Experiment Settings}
We build a complex-valued neural network from scratch. The derivation of back-propagation in complex-valued neural network can be found in Appendix. \ref{app:2}. MNIST is our dataset and gradient descent is applied to optimize the network \citep{deng2012mnist}. Experiments can be run on a personal computer with pipeline as follows. In different trails, while training neural the network, we prune most of the information from the weights except a certain information type after each step of gradient descent. Candidate types of information to preserve are real part, imaginary part, amplitude and phase. Then the pruned model is directly used to predict on testing set. By comparing the test accuracy, we are able to examine the importance of the preserved type of information in the system. To guarantee the consistency, our benchmark test accuracy for comparison is obtained by randomly pruning half of the weights, i.e. masking them with zero, and we report its average test accuracy over $10$ continuous steps. 

It's worth mentioning that we pre-processed the input image by adding an extra phase angle to each real-valued pixel to enhance the stability of our training and improve the performance of the system, catering for our complex-valued neural network. The phase added is proportional to the magnitude of the original pixel. This heuristic pre-processing procedure actually adds no extra information. One possible intuition behind this is to better utilize the phase-related subspace of the neural network and keep the fairness between amplitude and phase. In fact, mapping real input to the complex space is a widely used technique when dealing with complex-valued neural networks \cite{Savitha2011}.

\begin{figure}[htbp]
\centering
\includegraphics[width=\textwidth]{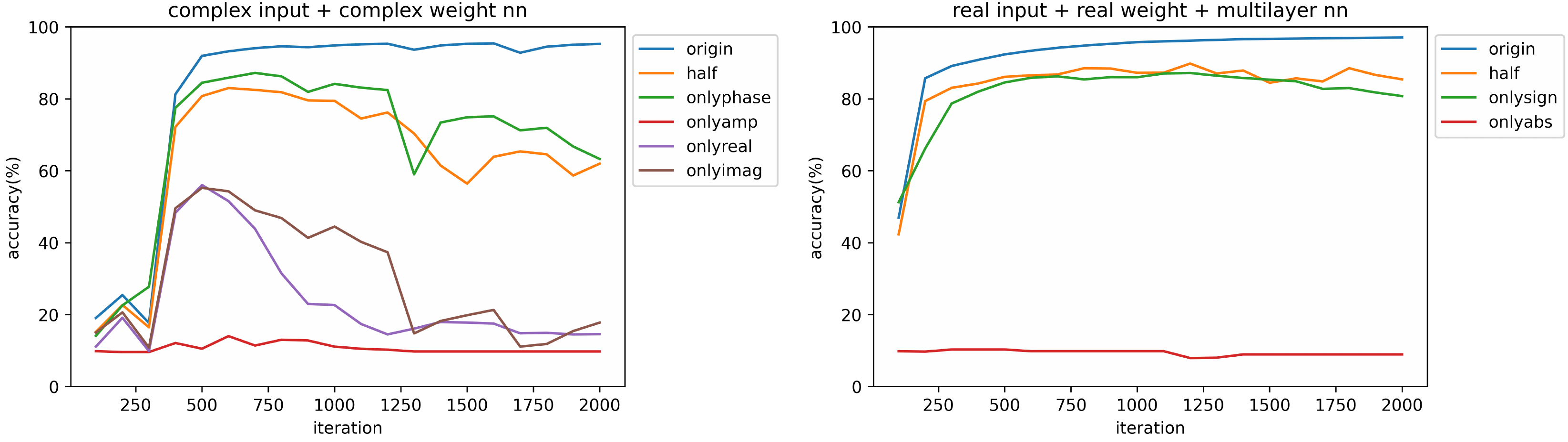}
\caption{\textbf{Left panel}: Results in complex-valued neural network with a hidden layer of width 100. The "origin" curve shows the test accuracy of the original model without pruning and the "half" curve corresponds to randomly pruning half of the weights. The curve "only$+$type" corresponds to the case where only the "type" information is preserved. \textbf{Right panel}: Results in real-valued 3-layer neural network which has 512 neurons in the first hidden layer and 100 neurons in the second hidden layer.}
\label{fig:1}
\end{figure}

\section{Results}
\label{complex}

\subsection{Complex-valued Neural Network}
\subsubsection{Main Results}
\label{main}
Our main result is shown on the left panel of Fig.\ref{fig:1}. Among all types of pruning, the phase curve (green) achieves the highest test accuracy, comparable to the half curve (orange), which is portrayed to be more powerful than other remaining accuracy curves. Especially, phase-preserved model outperforms amplitude-preserved model by a great margin. With only amplitude information, we can achieve no more than a random guess, which guarantees a $10\%$ accuracy. This result suggests that phase information matters far more than amplitude in complex-valued neural networks.

\begin{figure}[htbp]
\centering
\includegraphics[width=\textwidth]{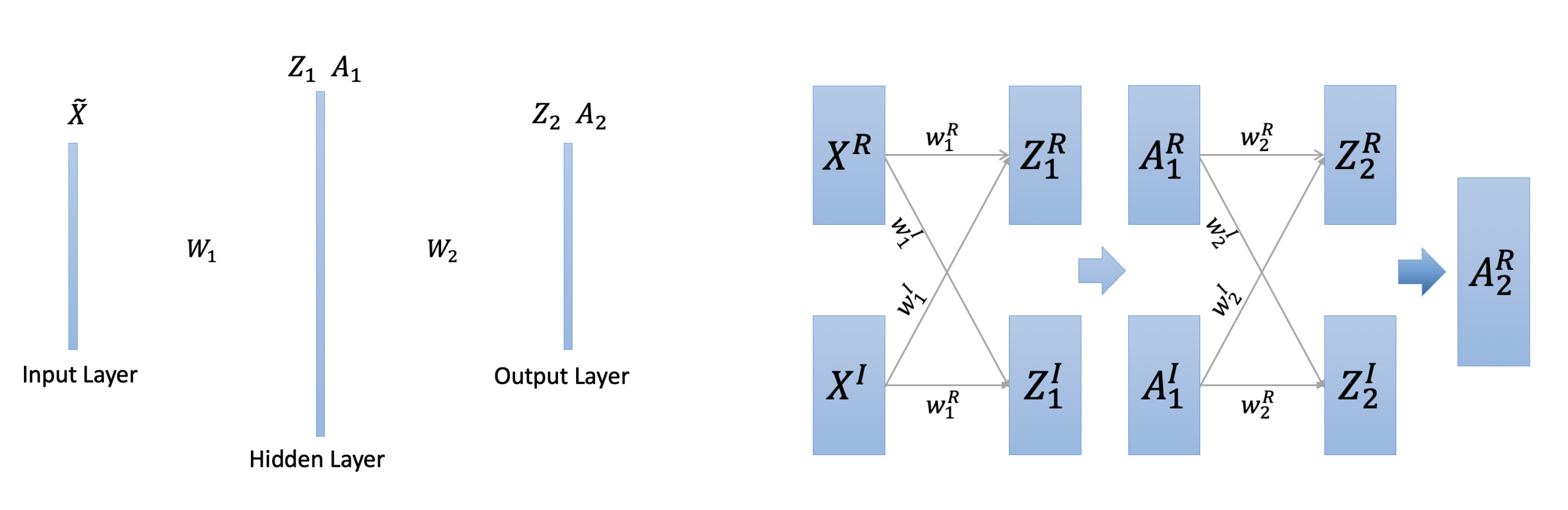}
\caption{\textbf{Left panel}: Complex-valued neural network with a hidden layer. \textbf{Right panel}: Decomposition of the network. The $I$ and $R$  superscript denotes the imaginary part and real part respectively.}
\label{fig:3}
\end{figure}

We also observe that the real and imaginary curve always keep comparable accuracy as the iteration goes. It seems natural since intuitively every complex number can be divided into its real and imaginary part, and both part store equal amount of information. A more explicit explanation is provided here. Essentially, motivated by the decomposition of complex number, we can decompose the whole network structure into its real part and imaginary part as Fig.\ref{fig:3} shows. By stacking these two component sub-networks, we end up with a neural network with doubled width, where both real and imaginary part of weights in the original network account for a half number of weights. However, a similar decomposition exists for phase-amplitude pair, while the phase curve performs much better than the amplitude curve. Is it a paradox? Absolutely not. Our explanation about decomposition only tells half of the story. What is exactly equal between the real and imaginary pair (or phase and amplitude pair) is the capacity of storing information, rather than the amount of information actually stored in both part, especially the amount of information that leads to the success of the original network. Therefore, it suggests a heterogeneity in information storage in complex-valued neural networks. The phase part indeed stores more useful information of the whole network even though its storage capacity is not distinguishing from its counterpart.

It's exciting to discover a clue to "phase matters more", but whether this heterogeneity holds robustly when varying the structure of network is unrevealed. We will verify our main result in wilder scenarios and meanwhile investigate how the depth and width effect our result in the following part. 

\subsubsection{Results with Different Widths and Depths}

\begin{figure}[htbp]
\centering
\includegraphics[width=\textwidth]{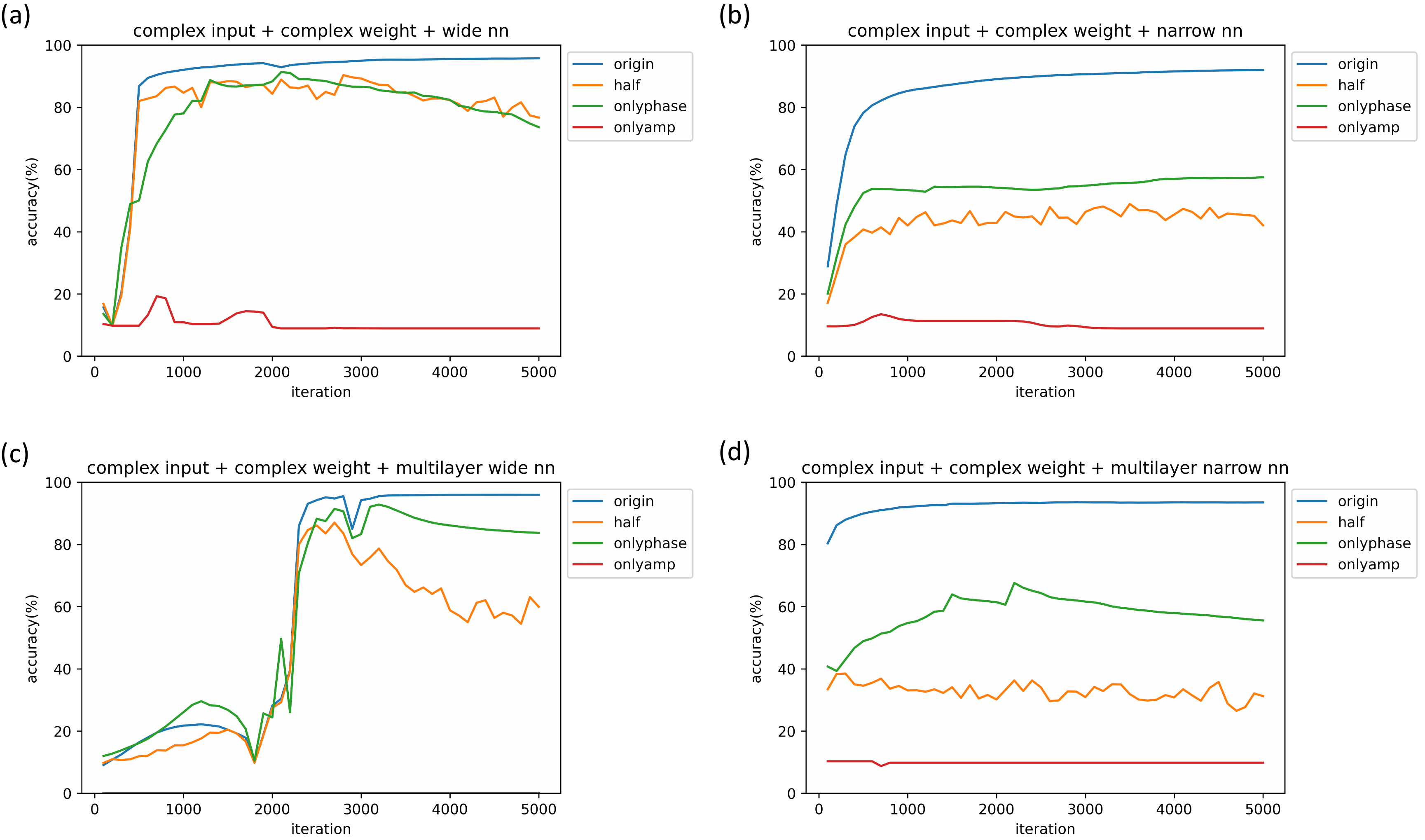}
\caption{Results in complex-valued neural networks with different widths and depths. \textbf{(a)} One hidden layer with 512 neurons. \textbf{(b)}  One hidden layer with 20 neurons. \textbf{(c)}  Two hidden layers with 512 neurons for the first and 100 neurons for the second. \textbf{(d)}  Two hidden layers with 50 neurons for the first and 20 neurons for the second.}
\label{fig:2}
\end{figure}

We first consider the impact of the width of network by varying the width of the hidden layer. Results are shown in the upper row of Fig.\ref{fig:2}. It is obvious that phase matters more than amplitude in both wide and narrow network cases. An extra observation is the phase curve outperforms the randomly-half-pruned curve by a considerable margin in the narrow neural networks, although the accuracy is relatively low because of the limited width of the hidden layer. 

We then consider the impact of the depth of the network. We add one more hidden layer to both the wide and narrow neural networks and list the corresponding results in the second row of Fig.\ref{fig:2}. By making a latitudinal comparison, we can see that in both cases of wide and narrow networks, the heterogeneity between phase and amplitude is enlarged when adding extra hidden layer, which suggests the possibility to leverage this heterogeneity in deep neural networks.

\subsection{Real-valued Neural Network}
\label{sec:real}

The importance of phase over amplitude in complex-valued neural network has been revealed in the previous section. In this section, we will extend the aforementioned heterogeneity to real-valued neural network.

The scheme for extension is straightforward. Since the sign and the absolute value of a real number can be interpreted as the counterpart of the phase and the amplitude of a complex number, that is, $e^{i\pi}=-1$ and $e^{2i\pi}=1$, pruning phase or amplitude in real-valued neural network is actually pruning the sign or absolute value of weights. Here pruning the absolute value means we normalize the absolute value of weights to $1$. It sounds bold if still phase matters more: it means even with binary weights, real-valued neural network retains prediction accuracy! 

The right panel of Fig.\ref{fig:1} shows the empirical results in a real neural network with two hidden layers. What the results show is super astonishing: phase still matters more! While the magnitude-preserved red curve still performs poorly, the performance of the only-sign curve is completely as competitive as that of the only-phase curve in all complex valued experiments, even though in real-valued case the binary sign space is much smaller than the phase space in complex-valued case. The model space for a complex-valued neural network can be divided equally into the phase space and the amplitude space, however, the division into the sign and the magnitude space for a real neural network is extremely biased. It suggests a stronger heterogeneity: in real-valued neural network, the "majority" of information is encoded in the "tiny" sign space while the "minority" of information is stored in the "vast" absolute value space.

\section{Conclusion and Discussion}
\label{sec:conclusion}

In this paper, we find that phase is more important than amplitude in complex-valued neural networks in the sense that preserving the phase information in weights does a better job in keeping prediction accuracy after pruning than preserving other types of information. Despite that widths and depths will influence the accuracy, "phase matters more" holds robustly in various neural network settings. This observation can be further extended to real-valued neural networks where the sign takes the place of the phase.

Finding the importance of phase brings us great potential in future work. The most straightforward insight is to leverage the phase information more in empirical work. For instance, it can enrich the techniques of network pruning, since the state-of-the-art pruning methods require ordering gradients of weight parameters in each step thus might be inefficient. Our pruning method used in experiments suggests its potential for fast pruning. Generalization to other neural networks such as CNN remains for future work. Also, if neural networks with binary weights (when we only care about the signals) can replace some parts of a large deep learning model without losing much accuracy, then optimization in logic computation, either in hardware layer or in algorithm layer, would be imaginable, where the computational cost and the memory occupation would be significantly reduced. Our work also provides an interesting view of theoretical interest, that is why this heterogeneity exists, which we haven't analyzed much in this paper but can be a good further step.

\section*{Broader Impact}

Network pruning has very broad applications in many field, and complex-valued neural network is always a topic of interest for decades. However, few connections have been built between these two. This work provides an interesting view of understanding neural networks with the insight from optical physics. It offers a possible method for analyzing the heterogeneity in deep learning models via network pruning. The authors do not observe any negative ethical impact about this work.

\section*{Acknowledgement}
The authors acknowledge the helpful discussion with Prof. Boris Hanin and the suggestions from anonymous reviewers.

\medskip
\bibliographystyle{unsrtnat}
\bibliography{neurips_2021}

\appendix

\section{Appendix}

\subsection{Neural networks and Wave optics}
\label{app:1}

\begin{figure}[htbp]
\centering 
\subfigure[4F correlator in wave optics. The very left plane is the object plane and the imaging plane is on the very right. Planes in the middle are wavefront-transformation planes, which are usually optical devices such as lenses in reality. This figure is credited to \url{https://en.wikipedia.org/wiki/Fourier_optics}.]{ 
\label{fig:subfig:a0} 
\includegraphics[scale=0.4]{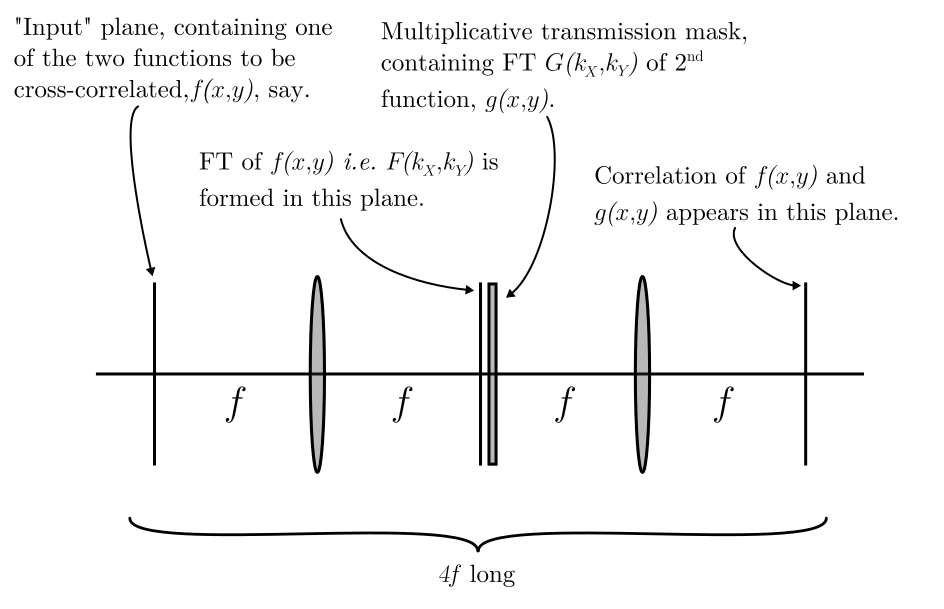}} 
\subfigure[Reconstruct figure from its Fourier transform. The upper right subfigure shows the case where only magnitude is preserved to reconstruct, while the lower right one corresponds to the case with only phase. We credit this figure to \textit{Igor Aizenberg, Manhattan College}. ]{ 
\label{fig:subfig:b0} 
\includegraphics[scale=0.4]{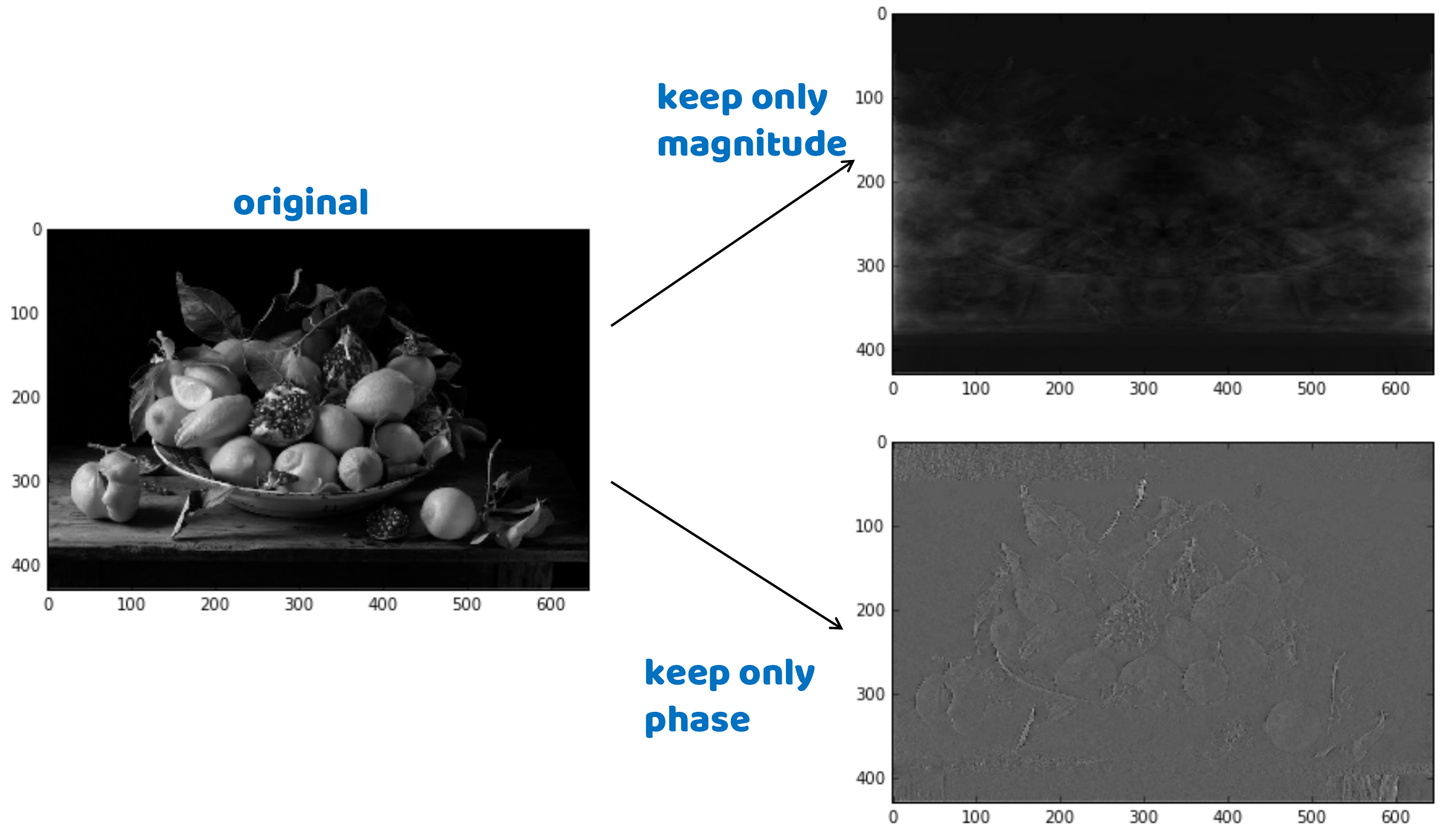}} 
\caption{Intuitions from perspective in Physics} 
\label{fig:p+m} 
\end{figure}

It's a common intuition in physics that complex methods always bring non-trivial properties to the system. Thus we can find a clue to the non-trivial property lies in complex-valued neural networks by viewing the network from a physical perspective. In Fig.\ref{fig:subfig:a0}, the object plane and imaging plane are analogous to the input and output layer in neural networks respectively. The wave propagation in air is essentially the same as  the multiplication by weights in neural network since it can also be converted to matrix transformation. Besides, wavefront-transformation on planes is the counterpart of activation function. In conclusion, the wave propagation through a whole imaging systems is an analogy to the forward propagation in a neural network, which is more precisely, the complex-valued neural network, since physicists always describe optical wave by complex amplitude, and the diffraction integration equation in physics is also an complex-valued formula. 

\subsection{Backward-propagation Algorithm for Complex-valued Neural Networks}
\label{app:2}

In complex-valued neural network, we cannot simply replace the real values by complex values. If we run the code in this way, the program will break up because of the non-intuitive analytical properties of the complex algebra \cite{Simone2018}. So we need to re-analyze the backward-propagation which is compatible with complex-valued neural networks \cite{popa2017}.

We list some of the notations that we will frequently use later.

\noindent\begin{longtable}{@{}l @{\quad \quad} l@{}}
$\widetilde{\mathbf{X}}=(\mathbf{X},b)\in\mathbb{C}^{(n+1)\times m}$  & Input MNIST data with n dimensions and m numbers\\
$\mathbf{W}_1\in\mathbb{C}^{h\times (n+1)}$  & Weight matrix connecting input layer and hidden layer\\
$\mathbf{Z}_1=\mathbf{W}_1\widetilde{\mathbf{X}}\in\mathbb{C}^{h\times m}$  & Values of nodes on hidden layer before activation\\
$\mathbf{A}_1=f_1(\mathbf{Z}_1)\in\mathbb{C}^{h\times m}$  & Values of nodes on hidden layer. $f_1$: activation function\\
$\mathbf{W}_2\in\mathbb{C}^{c\times h}$  & Weight matrix connecting hidden layer and output layer\\
$\mathbf{Z}_2=\mathbf{W}_2\mathbf{A}_1\in\mathbb{C}^{c\times m}$  & Values of nodes on output layer before activation\\
$\mathbf{A}_2=f_2(\mathbf{Z}_2)\in\mathbb{C}^{c\times m}$  & Values of nodes on output layer. $f_2$: activation function\\
$\mathbf{Y}\in\mathbb{C}^{c\times m}$  & One-hot encoded labels\\
\end{longtable}

The sketch of our network with these notations is provided in Fig.\ref{fig:1}. Specifically, in our system, we will use $f_1(x) = \mathbb{C}ReLU (x) = ReLU(x.real) + i*ReLU(x.imag)$ and  $f_2(z) = modSoftmax (z) = \frac{e^{|z|^2}}{\mathbf{1}^Te^{|z|^2}}$ as our activation functions. Since the input MNIST data is normalized to the range $[0,1]$, we add an extra phase in the range $[0,\pi]$ which is proportional to its magnitude to make the best of the complex system. 

To optimizing the weights $\mathbf{W}_1$ and $\mathbf{W}_2$, we need to calculate their gradient $\frac{dL}{d\mathbf{W}_1}$ and $\frac{dL}{d\mathbf{W}_2}$. By chain rule:

\begin{equation}\nonumber
\frac{dL}{d\mathbf{W}_2} = \frac{dL}{d\mathbf{A}_2}\frac{d\mathbf{A}_2}{d\mathbf{Z}_2}\frac{d\mathbf{Z}_2}{d\mathbf{W}_2}
\end{equation}

Here L is the loss function: $L=\frac{1}{m}\sum H$ and H is the cross entropy $H=-\sum p \mathrm{log} q = -\sum Y \mathrm{log} \mathbf{A}_2$. Thus it is easy to get:

\begin{align*}
\frac{dL}{d\mathbf{Z}_2} & = \frac{dL}{d\mathbf{A}_2}\frac{d\mathbf{A}_2}{d\mathbf{Z}_2}\\
& = \frac{dL}{d\mathbf{A}_2}\left.\dfrac{df_2'(z)}{dz}\right|_{z=|\mathbf{Z}_2|^2}\frac{d|\mathbf{Z}_2|^2}{d\mathbf{Z}_2}\\
& = (\mathbf{A}_2-\mathbf{Y}) \otimes 2\mathbf{Z}_2
\end{align*}

Here $\otimes$ denotes the element-wise multiplication, and $f_2'(z) = Softmax (z) = \frac{e^{z}}{\mathbf{1}^Te^{z}}$ is the original Softmax function. We use $\mathbf{Z}_2^R$ to represent its real part, and $\mathbf{Z}_2^I$ for its imaginary part. Since $L, \mathbf{A}_2,\mathbf{Y}$ are all real numbers, we have: 

$$\frac{dL}{d\mathbf{Z}_2^R} = (\mathbf{A}_2-\mathbf{Y}) \otimes 2\mathbf{Z}_2^R$$
$$\frac{dL}{d\mathbf{Z}_2^I} = (\mathbf{A}_2-\mathbf{Y}) \otimes 2\mathbf{Z}_2^I$$

Then we get our equation for obtaining $\frac{dL}{d\mathbf{W}_2}$:

\begin{equation}\nonumber
\frac{dL}{d\mathbf{W}_2^R} = \frac{dL}{d\mathbf{Z}_2^R}\frac{d\mathbf{Z}_2^R}{d\mathbf{W}_2^R} + \frac{dL}{d\mathbf{Z}_2^I}\frac{d\mathbf{Z}_2^I}{d\mathbf{W}_2^R}
\end{equation}
\begin{equation}\nonumber
\frac{dL}{d\mathbf{W}_2^I} = \frac{dL}{d\mathbf{Z}_2^R}\frac{d\mathbf{Z}_2^R}{d\mathbf{W}_2^I} + \frac{dL}{d\mathbf{Z}_2^I}\frac{d\mathbf{Z}_2^I}{d\mathbf{W}_2^I}
\end{equation}

$\mathbf{Z}_2$ can be expanded as: $\mathbf{Z}_2 = \mathbf{W}_2 \mathbf{A}_1 = (\mathbf{W}_2^R \mathbf{A}_1^R-\mathbf{W}_2^I \mathbf{A}_1^I)+i(\mathbf{W}_2^R \mathbf{A}_1^I+\mathbf{W}_2^I \mathbf{A}_1^R)$, thus we have: $\frac{d\mathbf{Z}_2^R}{d\mathbf{W}_2^R} = {\mathbf{A}_1^T}^R$, $\frac{d\mathbf{Z}_2^I}{d\mathbf{W}_2^R} = {\mathbf{A}_1^T}^I$, $\frac{d\mathbf{Z}_2^R}{d\mathbf{W}_2^I} = -{\mathbf{A}_1^T}^I$, $\frac{d\mathbf{Z}_2^I}{d\mathbf{W}_2^I} = {\mathbf{A}_1^T}^R$. Substituting the results back we get:

\begin{equation}\label{eq:W2}
\begin{split}
\frac{dL}{d\mathbf{W}_2^R} = \{(\mathbf{A}_2-\mathbf{Y}) \otimes 2\mathbf{Z}_2^R\} {\mathbf{A}_1^T}^R + \{(\mathbf{A}_2-\mathbf{Y}) \otimes 2\mathbf{Z}_2^I\} {\mathbf{A}_1^T}^I\\
\frac{dL}{d\mathbf{W}_2^I} = \{(\mathbf{A}_2-\mathbf{Y}) \otimes 2\mathbf{Z}_2^I\} {\mathbf{A}_1^T}^R - \{(\mathbf{A}_2-\mathbf{Y}) \otimes 2\mathbf{Z}_2^R\} {\mathbf{A}_1^T}^I
\end{split}
\end{equation}

Then we calculate $\frac{dL}{d\mathbf{A}_1}$. Since we have

\begin{equation}\nonumber
\frac{dL}{d\mathbf{A}_1^R} = \frac{dL}{d\mathbf{Z}_2^R}\frac{d\mathbf{Z}_2^R}{d\mathbf{A}_1^R} + \frac{dL}{d\mathbf{Z}_2^I}\frac{d\mathbf{Z}_2^I}{d\mathbf{A}_1^R}
\end{equation}
\begin{equation}\nonumber
\frac{dL}{d\mathbf{A}_1^I} = \frac{dL}{d\mathbf{Z}_2^I}\frac{d\mathbf{Z}_2^I}{d\mathbf{A}_1^I} + \frac{dL}{d\mathbf{Z}_2^R}\frac{d\mathbf{Z}_2^R}{d\mathbf{A}_1^I}
\end{equation}

Following similar steps and using considering the dimension compatibility, we have:

\begin{equation}\nonumber
\begin{split}
\frac{dL}{d\mathbf{A}_1^R} ={\mathbf{W}_2^T}^R \{(\mathbf{A}_2-\mathbf{Y}) \otimes 2\mathbf{Z}_2^R\}  + {\mathbf{W}_2^T}^I \{(\mathbf{A}_2-\mathbf{Y}) \otimes 2\mathbf{Z}_2^I\}\\
\frac{dL}{d\mathbf{A}_1^I} ={\mathbf{W}_2^T}^R \{(\mathbf{A}_2-\mathbf{Y}) \otimes 2\mathbf{Z}_2^I\}  - {\mathbf{W}_2^T}^I \{(\mathbf{A}_2-\mathbf{Y}) \otimes 2\mathbf{Z}_2^R\}
\end{split}
\end{equation}

By using this formula we are able to calculate $\frac{dL}{d\mathbf{Z}_1}$. The function $f_1(x) = \mathbb{C}ReLU (x) = ReLU(x.real) + i*ReLU(x.imag)$ has very good property in derivatives: $\frac{d\mathbf{A}_1^R}{d\mathbf{Z}_1^R}=\frac{1}{2}(1+\mathrm{sign}(\mathbf{Z}_1^R))$, $\frac{d\mathbf{A}_1^I}{d\mathbf{Z}_1^I}=\frac{1}{2}(1+\mathrm{sign}(\mathbf{Z}_1^I))$, and $\frac{d\mathbf{A}_1^I}{d\mathbf{Z}_1^R}=\frac{d\mathbf{A}_1^R}{d\mathbf{Z}_1^I}=0$. Thus we can obtain:

\begin{equation}\nonumber
\begin{split}
\frac{dL}{d\mathbf{Z}_1^R} =\frac{dL}{d\mathbf{A}_1^R} \otimes \frac{1}{2}(1+\mathrm{sign}(\mathbf{Z}_1^R))\\
\frac{dL}{d\mathbf{Z}_1^I} =\frac{dL}{d\mathbf{A}_1^I} \otimes \frac{1}{2}(1+\mathrm{sign}(\mathbf{Z}_1^I))
\end{split}
\end{equation}

And $\mathbf{Z}_1$ can also be expanded as: $\mathbf{Z}_1 = \mathbf{W}_1 \widetilde{\mathbf{X}} = (\mathbf{W}_1^R \widetilde{\mathbf{X}}^R - \mathbf{W}_1^I \widetilde{\mathbf{X}}^I) + i(\mathbf{W}_1^I \widetilde{\mathbf{X}}^R+\mathbf{W}_1^R \widetilde{\mathbf{X}}^I)$, so finally we will get:

\begin{equation}\label{eq:W1}
\begin{split}
\frac{dL}{d\mathbf{W}_1^R} = \frac{dL}{d\mathbf{Z}_1^R} (\widetilde{\mathbf{X}}^T)^R+\frac{dL}{d\mathbf{Z}_1^I} (\widetilde{\mathbf{X}}^T)^I\\
\frac{dL}{d\mathbf{W}_1^I} = \frac{dL}{d\mathbf{Z}_1^I}(\widetilde{\mathbf{X}}^T)^R-\frac{dL}{d\mathbf{Z}_1^R}(\widetilde{\mathbf{X}}^T)^I
\end{split}
\end{equation}

The equations (\ref{eq:W2}) and (\ref{eq:W1}) are the formulas that we will use in our program to update the parameters $\mathbf{W}_1$ and $\mathbf{W}_2$ when training the complex-valued neural networks.

\end{document}